\documentclass[letterpaper]{article} 
\usepackage{aaai2026}  
\usepackage{times}  
\usepackage{helvet}  
\usepackage{courier}  
\usepackage[hyphens]{url}  
\usepackage{graphicx} 
\usepackage{natbib}  
\usepackage{caption} 
\usepackage{algorithm}

\usepackage{newfloat}
\usepackage{listings}
\DeclareCaptionStyle{ruled}{labelfont=normalfont,labelsep=colon,strut=off} 
\floatstyle{ruled}
\newfloat{listing}{tb}{lst}{}
\floatname{listing}{Listing}
\title{Run, Ruminate, and Regulate: A Dual-process Thinking System for Vision-and-Language Navigation}
\author{
    Yu Zhong\textsuperscript{\rm 1,2}, Zihao Zhang\textsuperscript{\rm 1,3}\thanks{corresponding authors}, Rui Zhang\textsuperscript{\rm 1}, Lingdong Huang\textsuperscript{\rm 1,2}, Haihan Gao\textsuperscript{\rm 1,4}, Shuo Wang\textsuperscript{\rm 1,2}, \\Da Li\textsuperscript{\rm 5,2}, Ruijian Han\textsuperscript{\rm 6}, Jiaming Guo\textsuperscript{\rm 1}, Shaohui Peng\textsuperscript{\rm 7}, Di Huang\textsuperscript{\rm 1}, \\Yunji Chen\textsuperscript{\rm 1,2}\footnotemark[1]
}
\affiliations{
    \textsuperscript{\rm 1}State Key Lab of Processors, Institute of Computing Technology, Chinese Academy of Sciences\\
    \textsuperscript{\rm 2}University of Chinese Academy of Sciences (UCAS)\\
    \textsuperscript{\rm 3}Institute of AI for Industries (IAII), Chinese Academy of Sciences\\
    \textsuperscript{\rm 4}University of Science and Technology of China\\
    \textsuperscript{\rm 5}Institute of Computing Technology, Chinese Academy of Sciences\\
    \textsuperscript{\rm 6}Department of Data Science and Artificial Intelligence, The Hong Kong Polytechnic University\\
    \textsuperscript{\rm 7}Institute of Software, Chinese Academy of Sciences
}

\usepackage{tcolorbox}
\usepackage{enumitem}
\usepackage{amssymb}
\usepackage{makecell}
\tcbuselibrary{skins}
\usepackage{graphicx}
\usepackage{subcaption}
\usepackage{amsmath}
\usepackage{algorithm}
\usepackage{algpseudocode}
\usepackage{multirow}
\usepackage[table]{xcolor}
\usepackage{caption}
\usepackage{booktabs}
\newcommand{\ZY}[1]{{\color[rgb]{0,0,0}#1}} 
\usepackage{bibentry}

\begin{document}
\maketitle
\begin{abstract}
Vision-and-Language Navigation (VLN) requires an agent to dynamically explore complex 3D environments following human instructions.
Recent research underscores the potential of harnessing large language models (LLMs) for VLN, given their commonsense knowledge and general reasoning capabilities.
Despite their strengths, a substantial gap in task completion performance persists between LLM-based approaches and domain experts, as LLMs inherently struggle to comprehend real-world spatial correlations precisely.
Additionally, introducing LLMs is accompanied with substantial computational
cost and inference latency.
To address these issues, we propose a novel dual-process thinking framework dubbed \textbf{R$^3$}, integrating LLMs’ generalization capabilities with VLN-specific expertise in a zero-shot manner.
The framework comprises three core modules: \textbf{R}unner, \textbf{R}uminator, and \textbf{R}egulator.
The Runner is a lightweight transformer-based expert model that ensures efficient and accurate navigation under regular circumstances.
The Ruminator employs a powerful multimodal LLM as the backbone and adopts chain-of-thought (CoT) prompting to elicit structured reasoning.
The Regulator monitors the navigation progress and controls the appropriate thinking mode according to three criteria, integrating Runner and Ruminator harmoniously.
Experimental results illustrate that R$^3$ significantly outperforms other state-of-the-art methods, exceeding 3.28\% and 3.30\% in SPL and RGSPL respectively on the REVERIE benchmark.
This pronounced enhancement highlights the effectiveness of our method in handling challenging VLN tasks.
\begin{links}
    \link{Code}{https://github.com/IAII-CAS/navigation_R3}
\end{links}
\end{abstract}
\begin{figure}[!t]
    \centering
    \includegraphics[width=.95\linewidth]{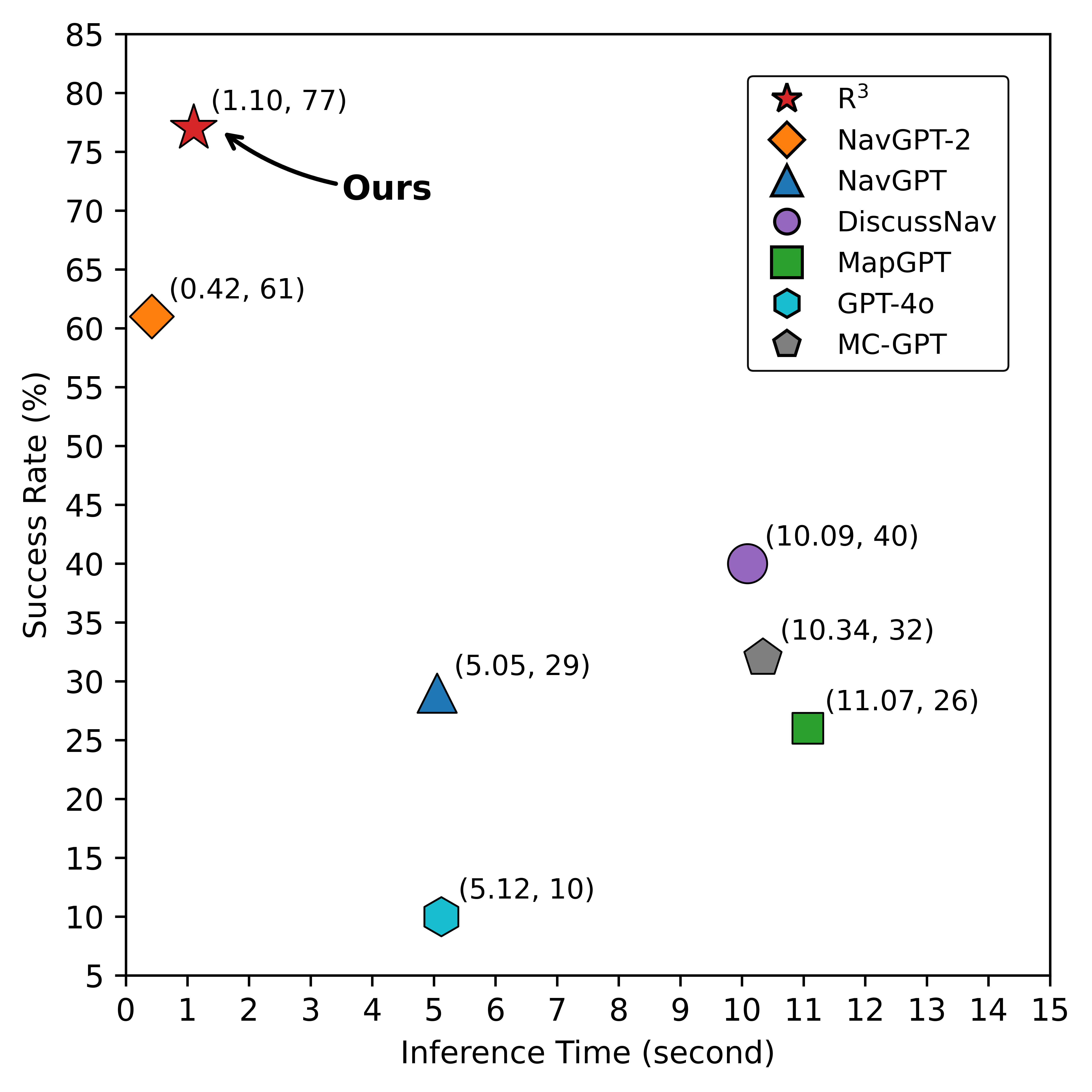}
    \caption{Comparison of inference efficiency and navigation performance. Our R$^3$ requires only one-fifth of the inference time compared with other LLM-assisted methods. NavGPT-2 exhibits a little better efficiency since it deploys LLMs locally while others query the GPT model via API.}
    \label{fig:efficiency}
\end{figure}

\begin{figure*}[!t]
    \centering
    \includegraphics[width=\linewidth]{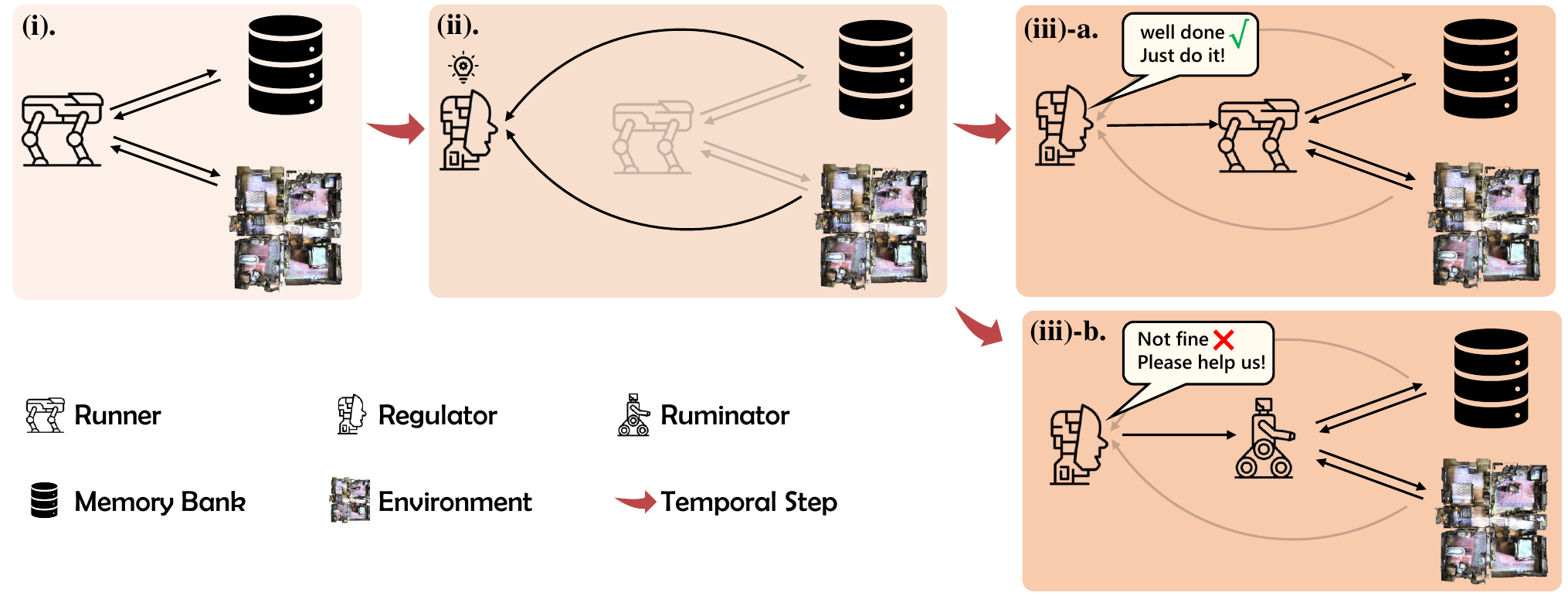}
    \caption{Overview of the proposed R$^3$.
    Our system comprises three core modules: Runner, Ruminator, and Regulator. The working flow operates as (i)$\rightarrow$(ii)$\rightarrow$(iii). The navigation initiates with the Runner. For each timestep, the Regulator evaluates the current condition. If the condition is nominal, the Runner proceeds ((iii)-a); otherwise, the Ruminator engages to resolve exceptions ((iii)-b)}
    \label{fig:teaser}
\end{figure*}

\section{Introduction}
Vision-and-language navigation (VLN) requires an embodied agent to adhere to human directives, perceive visual surroundings, and navigate through photorealistic environments to reach the target location \citep{R2R}.
This exploratory interaction form is on the cusp of embodied intelligence research due to its potential significance for service robotics, enabling these machines to automatically move toward designated areas and perform downstream tasks.
Despite notable advancements achieved, one key hindrance to real-world applications concerns agents' limited generalizability to new scenarios.

With the considerable development of large language models (LLMs), emerging studies have sought to address the generalizing issue by incorporating LLMs.
This stems primarily from LLMs' rich commonsense knowledge and powerful reasoning abilities, which facilitate enhanced cross-modal understanding and long-term planning across \ZY{diverse} environments.
Several works \citep{NavGPT-2, NaviLLM} adopt sequence-to-sequence-based behavior cloning (BC) to learn action patterns from navigation oracles through large-scale instruction-trajectory pairs.
Specifically, they transform visual observations into linguistic descriptions or encoded representations using pretrained vision–language models, which are then fed into LLMs. From the resulting textual output, subsequent actions can be extracted.
However, these methods still underperform specialist VLN models, primarily due to (1) the scarcity of VLN-specific training data for such large-scale backbones; and (2) the degradation of commonsense reasoning capabilities caused by tuning, which is crucial for demanding interactive tasks such as VLN.
In contrast, another line of approaches \citep{NavGPT, DiscussNav} utilizes more powerful proprietary LLMs such as GPT-4 to determine actions in a zero-shot manner, thereby avoiding defects caused by tuning.
Crafted prompts can potentially elicit more comprehensive perceptual summaries and deliberate reasoning to inform decision-making, thereby heightening the likelihood of locating the target.
Carefully designed prompts can elicit richer perceptual summaries and multi-step reasoning to inform decision making, thereby increasing the likelihood of locating the target.
However, most existing zero-shot methods fully delegate the navigation to LLMs.
Despite excelling at commonsense reasoning and high-level planning, LLMs' capacity to comprehend real-world spatial layouts and geometric structures has not been well studied, which may yield suboptimal decision-making when VLN-specific knowledge is absent.
In addition, the substantial inference latency of LLMs exacerbates the efficiency–accuracy imbalance, hindering the deployment of the VLN field where real-time responsiveness is commonly required.

Our goal is to devise a strategy that integrates the advantages of LLMs and domain experts, deriving maximum utility of LLMs' commonsense reasoning capabilities while incorporating in-context expertise.
\ZY{Inspired by the dual process theory \citep{kahneman2011thinking}, we propose R$^3$, a framework that emulates human cognition to tackle complex navigation tasks.}
Specifically, as illustrated in Fig.~\ref{fig:teaser}, our framework comprises three primary modules: \textbf{R}unner, \textbf{R}uminator, and \textbf{R}egulator.
The Runner module employs fast and intuitive thinking for \ZY{routine scenarios}, whose architecture is built upon a lightweight, transformer-based VLN expert.
The Ruminator module simulates slow and methodical thought, proposed for handling anomalous scenarios.
We introduce the multimodal LLM GPT-4 as the backbone and adopt chain-of-thought (CoT), an effective in-context learning technique, to engage multi-step reasoning.
The Regulator module is responsible for adaptively evaluating the current navigational situation and controlling the module switching.
To achieve this, we design a sophisticated two-stage switching mechanism, relying on three criteria tailored to VLN: \textit{looping}, \textit{scoring}, and \textit{ending}.
In addition, the Regulator employs a critical formulation process dedicated to clearing out unnecessary history, facilitating more effective engagement by the Ruminator.

We measure our approach on two categories of VLN benchmarks: \textit{fine-grained navigation} (R2R) and \textit{coarse-grained navigation} (REVERIE \citep{Reverie}).
Experimental results demonstrate that R$^3$ significantly outperforms state-of-the-art methods.
It is worth mentioning that R$^3$ is especially effective in handling complex tasks such as REVERIE, surpassing others by more than 3.28 and 3.30 points in SPL and RGSPL, respectively.
Furthermore, as depicted in Fig.~\ref{fig:efficiency}, R$^3$ is more time-efficient, requiring significantly less inference time per action than other LLM-based methods.
Superior performance verifies both the effectiveness and efficiency of our proposed R$^3$ framework in the VLN context.

\section{Related work}
\subsection{Vision-and-Language Navigation (VLN)}
In the VLN task, an embodied agent is required to navigate to a target location following natural instructions. 
Early works investigate cross-modal attention to enhance text-vision grounding and extract goal-relevant visual representations under fine-grained supervision \citep{Lxmert, VLN_BERT}.
A series of subsequent studies \citep{HAMT, DUET, BEVBert, GridMM} advances compelling methodologies by highlighting the essentials of topological maps to aggregate historical representations and facilitate long-term planning.
To gain better generalizability in unseen environments, pretraining \citep{HOP, HOP+}, data augmentation \citep{FDA, Envedit, ScaleVLN, PanoGen, RoomTour3D, Wang_2025}, commonsense knowledge incorporation \citep{KERM, mohammadi2024augmented}, reinforcement learning \citep{bundele2024scaling}, test-time adaptation \citep{fstta}, and other online learning techniques have been widely explored for VLN agents.
Some works \citep{NavCoT, pan2024langnav} attempt to finetune an open-source LLM into a VLN generalist.
\ZY{However, the generalist reasoning abilities of LLMs can be compromised when grounded with respect to a specific task.}
Some recent research \citep{liang-etal-2024-cornav} construct a linguisticly formed navigation agent that prompts the LLMs with the instruction and textually represented observations to determine actions in a zero-shot manner.
\citep{MapGPT} takes a step further by building an online topological map to store node information and activate global exploration. 
\subsection{Dual-Process Theory}
Stemming from neuroscience, the dual-process theory delineates the processing mechanisms of human cognition as a complex collaboration between two distinguished cognitive systems: the fast thinking system, which enables swift, automatic responses to real-time sensory information, and the slow thinking system, which excels in methodical analysis and deliberate reasoning, responsible for complicated tasks or high-level decision-making.
By incorporating the complementary strengths of two systems, dual-process-based models are capable of solving complex tasks both effectively and efficiently.
In recent years, the dual-process theory has demonstrated substantial potential across various challenging real-world applications, including autonomous driving \citep{ zhang2025chameleonfastslowneurosymboliclane}, embodied intelligence \citep{christakopoulou2024agentsthinkingfastslow}, and robotics \citep{wen2024vidman}.
The core focus of developing a dual-process thinking system lies in coordinating two systems to operate in concert, exploiting each system to its maximum potential.
In this work, we propose the first dual-process framework for the demanding task VLN.

\begin{figure*}[!t]
    \centering
    \includegraphics[width=\linewidth]{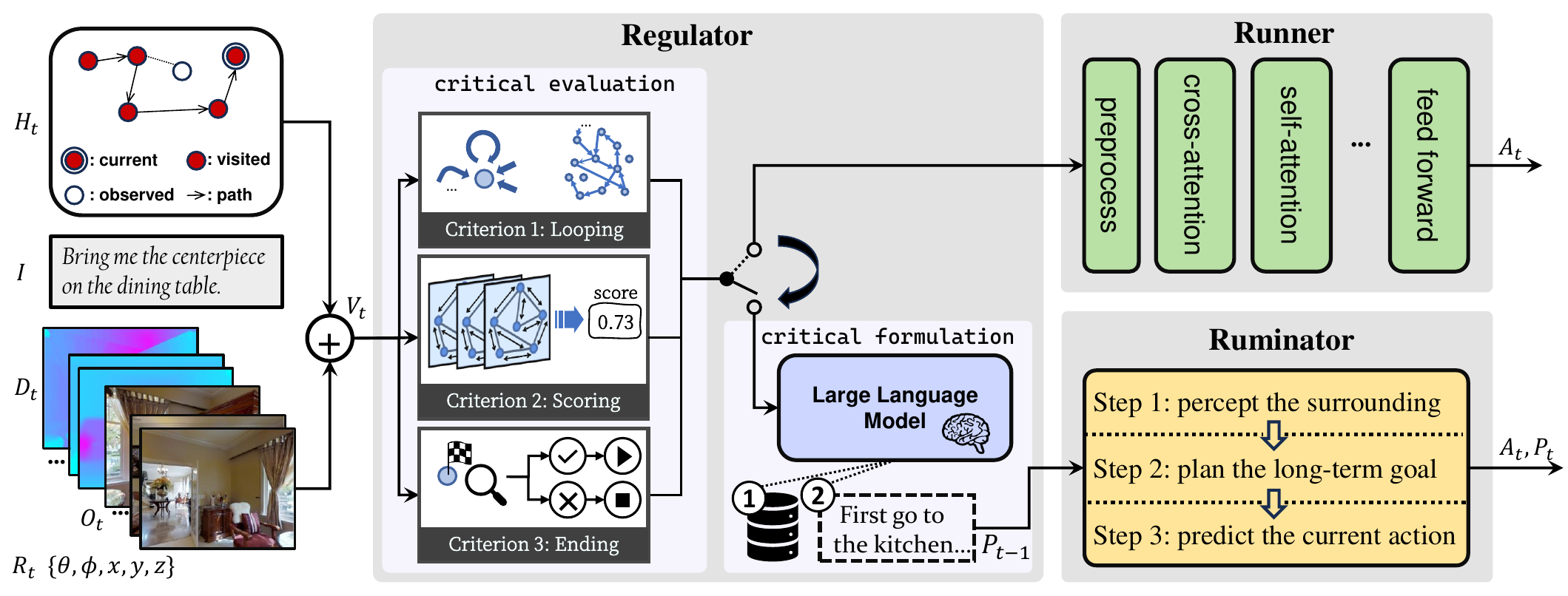}
    \caption{Overall pipeline of the proposed R$^3$.
    For each timestep $t$, the Regulator evaluates the navigation condition and switches to Ruminator if necessary, which resorts to mLLMs for resolving the anomalies; otherwise, the Runner, a lightweight, transformer-based VLN expert, proceeds with navigation efficiently.
    Here $V_t$ represents all inputs, including history $H_t$, instruction $I$, RGB-D images $O_t$, $D_t$, and pose $R_t$.}
    \label{fig_pipeline}
\end{figure*}

\begin{figure*}[!t]
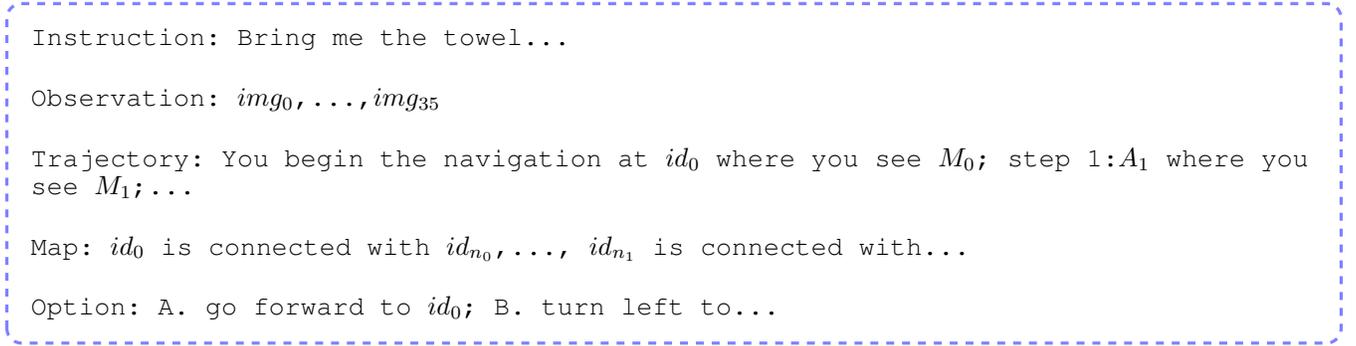
 
  \captionsetup{justification=raggedright,singlelinecheck=false}
  \begin{tcolorbox}[
      enhanced,
      width=\textwidth,
      colback=white,
      colframe=white,
      rounded corners,
      arc=5pt,
      left=2mm,
      right=2mm,
      boxsep=1mm,
      parbox=false,
      boxrule=1.2pt,
      borderline={1.2pt}{0pt}{blue!50,dashed},
      fontupper=\linespread{0.95}\selectfont, 
      before upper={\setlength{\parindent}{0pt}\noindent},
  ]
  \texttt{Instruction: Bring me the towel...}\\[2pt]

  \texttt{Observation: $img_{0}$,...,$img_{35}$}\\[2pt]

  \texttt{Trajectory: You begin the navigation at $id_{0}$ where you see $M_{0}$; step 1:$A_{1}$ where you see $M_{1}$;...}\\[2pt]

  \texttt{Map: $id_{0}$ is connected with $id_{n_{0}}$,..., $id_{n_{1}}$ is connected with...}\\[2pt]

  \texttt{Option: A. go forward to $id_{0}$; B. turn left to...}
  \end{tcolorbox}

  \caption{Textual template of inputs. By systematically formalizing the inputs, the Ruminator is capable of extracting current navigational circumstance in a more explicit and effective manner.}
  \label{fig:template}
\end{figure*}

\section{Methodology}
It is virtually impossible to handle all wayfinding circumstances through a single expert model due to its intrinsic inductive bias during training.
On the other hand, resorting to LLMs with commonsense reasoning abilities and wide knowledge coverage also suffers from significant domain gaps and inefficiency.
Accordingly, one simple and effective solution is to incorporate both strengths.
To achieve this, we propose the VLN framework $\text{R}^3$ based on the dual-process thinking, as depicted in Fig.~\ref{fig_pipeline}.
Specifically, our pipeline entails three functionally specialized modules: the Runner, which simulates the fast thinking system, the Ruminator, which plays the role of the slow thinking system, and the Regulator, responsible for monitoring the navigation progress and controlling the appropriate thinking mode accordingly.
The agent initiates an episode with the Runner.
For each timestep, the Regulator first evaluates the current condition and activates the Ruminator to engage when detecting anomalies.
Once switched, the Ruminator will take over the navigation exclusively without involving other modules until the episode ends.
As for nominal conditions, the Runner proceeds with the exploration.
By harmoniously coordinating these complementary modules, our approach is capable of yielding reliable navigation and computational efficiency across diverse navigation scenarios.

We begin this section by establishing the VLN problem.
Then we introduce the three modules of $R^3$ in the following subsections.

\subsection{Problem Formulation}
The VLN task is formulated as a partially observable Markov decision process, where an agent is required to follow a language instruction $I$ and navigate to the target destination by executing sequential actions with discrete time dynamics.
At each step $t$, the agent receives the real-time pose $R_t$ and an RGB panorama $O_t=\{o_t^i\}_{i=1}^{36}$ of surroundings from the environment $E$.
Each $o_t^i$ represents the perspective image with relative heading $\theta^i$ and elevation $\phi^i$ to the current orientation.
Among these, some perspective views are navigable, indicating the presence of adjacent viewpoints in these directions.
The agent needs to select one adjacent viewpoint as the action $A_t$.
To this end, the intrinsic goal of VLN is to optimize a policy $\pi$ with parameters $\Theta$ to predict the next action based on the instruction, history, and current observation: $\pi\bigl(A_t \mid I,\,O_t,\,H_t;\,\Theta)$.
Here the history includes all previous observations and actions: $H_t=\{O_0,A_0;...,O_{t-1},A_{t-1}\}$.
The aforementioned process continues until the agent predicts the $\texttt{[STOP]}$ action or exceeds the step limit.

\subsection{The Runner}
The Runner module operates as a reactive, fast-thinking component tailored to routine navigation scenarios.
We employ a lightweight transformer-based domain expert to construct the Runner.
Concretely, at timestep $t$, the Runner takes the RGB-D observations $O_t, D_t$, and the pose information $R_t$ to extract fine-grained features and store them into an egocentric grid memory with projection.
These features are then aggregated with instruction embeddings using a cross-modal transformer encoder and fed to a two-layer FFN for action prediction.
Benefiting from the grid-based topological memory, the agent is capable of facilitating long-term context awareness of environments.
To better utilize stored navigation dependencies, 
the Runner and the Ruminator share the memory bank.
Besides, the Runner encompasses around 160 M parameters only, guaranteeing rapid inference and on-time responsiveness.
However, Runner often yields suboptimal or even degraded behaviors when generalizing to some unseen scenarios.
This module is prone to making myopic decisions since its reactive policy is trained on limited distribution under a teacher-forcing scheme, leading to aimless wandering or repetitive actions when confronting unfamiliar situations.
Moreover, the Runner also struggles to fix mistakes once stepping into erroneous viewpoints since behavior cloning mimics the labeled trajectories rather than to learn the intrinsic navigation skills.

\subsection{The Ruminator}
To better handle troublesome cases that Runner may fail and further improve scalability, we present the Ruminator module for deliberate reasoning on resolving anomalies and realigning navigation with the intended objective from the perspective of commonsense knowledge.
We introduce GPT-4o as the navigation backbone of Ruminator and develop a CoT-based prompt system to activate the LLM's multi-step reasoning ability for corrective decision‐making.
Specifically, we structure the prompt system around three pivotal intermediate steps: perception, planning, and prediction. 
We first provide an input template as shown in Fig.~\ref{fig:template}.\\
\textbf{Perception}.
The first step enables the agent to perceive current surroundings and capture objects likely referenced by the instruction.
At time step $t$, the Ruminator receives the instruction $I$ and panoramic images $O_t$, prompting the LLM to convert these inputs into a fine-grained textual description of the environment, involving all important objects.\\
\textbf{Planning}.
After the perception step, we introduce the planning step to associate accumulated historical information and derive long-horizon planning for escaping critical scenarios.
We force the LLM to formulate the new planning $P_t$ relying on the given instruction $I$, previous planning $P_{t-1}$, textual descriptions acquired on the last step, and navigation history $H_t$.
Here $H_t$ consists of trajectory information and map information, as shown in the input template.
$id_k$ represents the ID of the $k$-th viewpoint visited by the agent.
$M_k$ represents the stored memory for the $k$-th viewpoint, which is the oriented perspective image when the agent is in the Runner state and instead the surrounding descriptions when in the Ruminator state.
$A_k$ comprises the taken action and the target destination.
The taken action is selected from $\{$\texttt{go forward to, turn left to, turn right to, turn back to}$\}$ according to the angle between the agent’s current orientation and the target.
And the target destination is the ID of the predicted viewpoint.\\
\textbf{Prediction}.
The Ruminator makes the final decision with the given instruction $I$, planning $P_t$, and $O_t'\subseteq O_t$ by choosing one action from candidate viewpoints. 
Here $O_t'$ contains all navigable perspective images.
The candidate options are explicitly listed, including the targeted direction and the candidate viewpoints, as demonstrated in Fig~\ref{fig:template}.
In the end, we update the history $H_{t}$ with selected action $A_t$ and neighboring viewpoints of the newly arrived location.
By leveraging our presented CoT strategy, the Ruminator can generate more interpretable planning and efficacious decisions for adeptly solving anomalous situations.

\definecolor{boldbg}{RGB}{214,233,202}
\definecolor{underbg}{RGB}{229,235,246}
\begin{table*}[!t]
  \centering
  \caption{Performance on R2R and Reverie datasets. The best results are in bold and highlighted in green, while the second are underlined and highlighted with blue. Our R$^3$ outperforms other methods on all metrics. Here \textbf{LLM-assisted} methods indicate that they adopt LLMs in a zero-shot manner but may require VLN-specific data in other places.}
  \label{tab:main}
  \resizebox{1\textwidth}{!}{
    \begin{tabular}{l|l|*{4}{c}|*{4}{c}}
      \toprule
      & \multirow{2}{*}{\centering{\textbf{Methods}}}
      & \multicolumn{4}{c|}{R2R Val Unseen}
      & \multicolumn{4}{c}{Reverie Val Unseen} \\
      \cmidrule(lr){3-6} \cmidrule(lr){7-10}
      &
      & TL & NE$\downarrow$ & SR$\uparrow$ & SPL$\uparrow$
      & SR$\uparrow$ & SPL$\uparrow$ & RGS$\uparrow$ & RGSPL$\uparrow$ \\
      \midrule
      \multirow{25}{*}{\centering\makecell{Behavior\\Cloning}} & Seq2Seq \citep{R2R} & 8.39  & 7.81   & 22     & -   & 4.20   & 2.84   & –      & 2.16 \\
        & RCM \citep{rcm}            & 11.46 & 6.09   & 43     & -   & 9.29   & 6.97   & –      & 3.89 \\
        & EnvDrop \citep{envdrop} & 10.70 & 5.22   & 52     & 48  & -      & -      & -      & -    \\
        & PREVALENT \citep{PRAVALENT}     & 10.19 & 4.71   & 58     & 53  & -      & -      & -      & -    \\
        & RecBERT \citep{VLN_BERT}       & 12.01 & 3.93   & 63     & 57  & 30.67  & 24.90  & 18.77  & 15.27 \\
        & HAMT \citep{HAMT}          & 11.46 & 3.65   & 66     & 61  & 32.95  & 30.20  & 18.92  & 17.28 \\
        & HOP \citep{HOP}           & 12.27 & 3.80   & 64     & 57  & 31.78  & 26.11  & 18.85 & 15.73 \\
        & DAVIS \citep{DAVIS}         & 12.65 & 3.16   & 67     & 61  & -      & -      & -      & -    \\
        & DSRG \citep{DSRG}          & -     & 3.00   & 73     & 62  & 47.83  & 34.02  & 32.69  & 23.37 \\
        & PanoGen \citep{PanoGen}       & 13.40 & 3.03   & 74     & 64  & -      & 33.44      & 32.80      & 22.45    \\
        & FDA \citep{FDA}           & 13.68 & 3.41   & 72     & 64  & 47.57  & 35.90  & 32.06  & 24.30 \\
        & KERM \citep{KERM}          & 13.54 & 3.22   & 72  & 61 & 49.02  & 34.83  & 33.97  & 24.14 \\
        & AZHP \citep{AZHP}          & 13.68 & 3.25   & 71     & 60  & 49.02  & 36.25  & 32.41  & 24.13 \\
        & CONSOLE \citep{CONSOLE}       & 13.59 & 3.00   & 73     & 63  & 50.07  & 34.40  & 34.05  & 23.33 \\
        & ESceme \citep{ESceme}        & 10.80 & 3.39   & 68     & 64  & -      & -      & -      & -    \\
        & SUSA \citep{SUSA}          & 12.18 & 3.06   & 73  & {\cellcolor{underbg}\underline{65}} & 51.75  & {\cellcolor{underbg}\underline{38.86}}  & {\cellcolor{underbg}\underline{35.02}}  & {\cellcolor{underbg}\underline{26.56}} \\
        & BEVBert \citep{BEVBert}       & 14.55 & {\cellcolor{underbg}\underline{2.81}}   & {\cellcolor{underbg}\underline{75}}     & 64  & {\cellcolor{underbg}\underline{51.78}}  & 36.37  & 34.71      & 24.44 \\
        & GridMM \citep{GridMM}        & 13.27     & 2.83      & {\cellcolor{underbg}\underline{75}} & 64  & 51.37  & 36.47  & 34.57   & 24.56 \\
        & DUET \citep{DUET}          & 13.94 & 3.31   & 72     & 60  & 46.98  & 33.73  & 32.15  & 23.03 \\
        & FAST \citep{Reverie}  & -     & -      & -      & -   & 14.40  & 7.19   & –      & 4.67 \\
        & SIA \citep{SIA}            & -     & -      & -      & -   & 31.53  & 16.28  & –      & 11.56 \\
        & Airbert \citep{Airbert}       & -     & -      & -      & -   & 27.89  & 21.88  & 18.23 & 14.18 \\
        & LANA \citep{LANA}          & 12.00     & -      & 68      & 62   & 48.31 & 33.86 & 32.86 & 22.77 \\
        & HOP+ \citep{HOP+}          & -     & -      & -      & -   & 36.07  & 31.13  & 22.49  & 19.33 \\
      \midrule
      \multirow{3}{*}{\centering\makecell{LLM\\fine-tuned}} & NavCoT \citep{NavCoT}        & 9.95  & 6.26   & 40     & 37  & 9.20   & 7.18  & -      & -    \\
        & NaviLLM \citep{NaviLLM}       & 12.81 & 3.51   & 67     & 59  & 28.10      & 21.04   & -      & -    \\   
        & NavGPT-2 \citep{NavGPT-2}      & 14.01 & 2.98   & 74     & 61  & -      & -      & -      & -    \\
      \midrule
      \multirow{6}{*}{\centering\makecell{LLM-\\assisted}} 
      & NavGPT \citep{NavGPT}        & 11.45 & 6.46   & 34     & 29  & 19.20 & 14.65   & -      & -    \\
        & MapGPT \citep{MapGPT}        & –     & 6.92   & 39     & 26  & 31.63      & 20.33      & -      & -    \\
        & DiscussNav \citep{DiscussNav}    & 9.69  & 5.32   & 43     & 40  & -      & -      & -      & -    \\
        & LangNav \citep{pan2024langnav}       & -     & 7.12      & 34     & 29   & -      & -      & -      & -    \\
        & MC-GPT \citep{mc-gpt}       & - & 7.76 & 22 & - & 19.43 & 9.65 & 8.86 & 5.14 \\
        & GPT-4 \citep{GPT4} & -     & 10.24      & 10     & 8   & -      & -      & -      & -    \\
      \cmidrule(lr){2-10}
      & R$^3$ (ours)   & 15.68  & {\cellcolor{boldbg}\textbf{2.76}}  & {\cellcolor{boldbg}\textbf{77}}  & {\cellcolor{boldbg}\textbf{66}}   & {\cellcolor{boldbg}\textbf{53.76}}  & {\cellcolor{boldbg}\textbf{42.14}}  & {\cellcolor{boldbg}\textbf{37.94}}  & {\cellcolor{boldbg}\textbf{29.86}} \\
      \bottomrule
    \end{tabular}
  }
\end{table*}

\subsection{The Regulator}
For our dual-process system, pinpointing appropriate moments for switching is of paramount importance.
As shown in Fig.~\ref{fig_pipeline}, we introduce the Regulator module in a hierarchical two-stage manner, focusing on \textbf{when} and \textbf{how} to switch respectively.
In the critical evaluation stage, we craft three complementary criteria to assess the agent’s current navigation state and determine whether intervention is required.
If positive, the agent switches from Runner to Ruminator and proceeds to the critical formulation stage.
Otherwise, it remains Runner and advances to the next timestep.
In the critical formulation stage, we employ GPT-4o to analyze the navigation progress and generate the corrective planning $P_{t}$ towards compensating for past deviations to ensure alignment with the intended objective.\\
\textbf{Stage 1: Critical Evaluation}. 
The Critical Evaluation stage determines \textbf{when} to switch from the Runner to the Ruminator. 
To achieve so, we design three criteria: \textit{looping}, \textit{scoring}, and \textit{ending} for the detection of potential anomalies.\\
\textit{Looping}. 
\ZY{One common failure mode is that the agent becomes trapped in cyclic traversal patterns due to navigational ambiguity.
We observe that frequent revisiting a certain viewpoint or excessive exploration often indicates that the agent is struggling to discover the intended path.
Inspired by these, we calibrate two thresholds $\tau_r,\tau_l\in \mathbb{R}$.
The Ruminator is triggered when the agent's maximum revisiting count across each viewpoint is larger than $\tau_r$ or the trajectory length exceeds $\tau_l$.
Recognizing looping is both simple and intuitive since all required information is documented in the history $H_t$.}\\
\textit{Scoring}.
While \textit{looping} can effectively address conspicuous failures, it struggles to capture the subtle anomalies entangled with historical context.
To solve these complex cases, we propose a scoring model based on Graph Neural Networks (GNN) to evaluate the likelihood of achieving navigation goals given the past trajectory.
Specifically, the scoring model utilizes two graph attention convolution layers with edge-encoding mechanisms to better extract graphic features from the historical map $H_t$, which is structured as a topology with nodes representing visited and observed viewpoints and edges representing paths.
To improve generalization and robustness without human annotations, we train the scoring model in a self-supervised manner by sampling trajectories and assigning algorithmically generated pseudo-labels.
We collect trajectories on the training and validation seen splits to avoid data leaking.
At every timestep $t$, we acquire the trajectory from $H_t$: $\mathcal{T}_t=\bigl(v_0, v_1, \dots, v_{t-1}\bigr)$, where $v_i$ represents the $i$-th visited viewpoint.
To enrich the topological representation, we take the position information, last visit timestep, and visual embedding together as the node feature for each viewpoint.
When a viewpoint remains unvisited, we approximate its visual embedding as the average of partial observations from neighboring visited viewpoints.
We then construct directed edges with every adjacent pair $<v_i, v_{i+1}>$ and complete the sampling.
As for labeling, we annotate each collected $\mathcal{T}_t$ with a binary label when an episode ends:
If the agent successfully reaches the destination or all viewpoints in $\mathcal{T}_t$ are included in the ground-truth path, we label $\mathcal{T}_t$ as $0$, indicating that this trajectory should be classified as nominal.
Otherwise we assign $1$.
During inference, we switch to the Ruminator if the output exceeds a threshold $\tau_g$.
The scoring criterion achieves fine-grained trajectory analysis by fusing structural connectivity patterns with semantic features, enabling early detection of deteriorating navigation states before catastrophic failure.\\
\textit{Ending}.
The Runner determines whether the episode is accomplished via a special token trained through behavior cloning (BC), which poses significant risks when generalizing to unseen situations.
To address this issue, we force the Regulator to examine whether the current location is the destination when the action $\texttt{[STOP]}$ is predicted by prompting GPT-4o with $I$ and $O_t$.
Benefiting from the powerful reasoning ability of LLMs, the ending criterion can prevent agents from ending episodes at the wrong location.\\
\textbf{Stage 2: Critical Formulation}
In some cases, the agent’s exploration may catastrophically deviate from the destination, rendering restarting from the start viewpoint preferable rather than continuing with the episode.
Thus, once the Ruminator engages, we additionally employ the critical formulation stage to evaluate the necessity of restarting from the initial state.
To faithfully simulate real-world deployment, we reset the memory bank whenever the agent restarts.
We prompt the LLM with the $I$, $H_t$, and $O_t$.
The critical formulation precludes the agent from accumulating misleading historical context of Runner, yielding enhanced accuracy through efficacious fixing strategies.

\section{Experiments}
\subsection{Setup}
\textbf{Datasets}. 
We evaluate our approach on two categories of VLN benchmarks: \textit{fine-grained navigation} (R2R) and \textit{coarse-grained navigation} (REVERIE).
Their visual environments are curated based on the Matterport3D dataset\citep{Matterport3D}, which includes 90 photo-realistic environments with 10,567 egocentric panoramas in total.
R2R is composed of 7,189 direct-to-goal shortest paths, each associated with 3 human-annotated navigation instructions.
The average length of an instruction is 29 words in R2R.
REVERIE includes 21,702 high-level instructions, requiring the agents to navigate and identify a remote object with ambiguous descriptions.
The instructions of REVERIE are ambiguous and only provide the target itself.
Their average length is 18 words, much shorter than those in R2R.
Therefore, REVERIE is deemed more challenging and closer to real-world robotic applications.
\begin{figure*}[t]
    \centering
    \includegraphics[width=\linewidth]{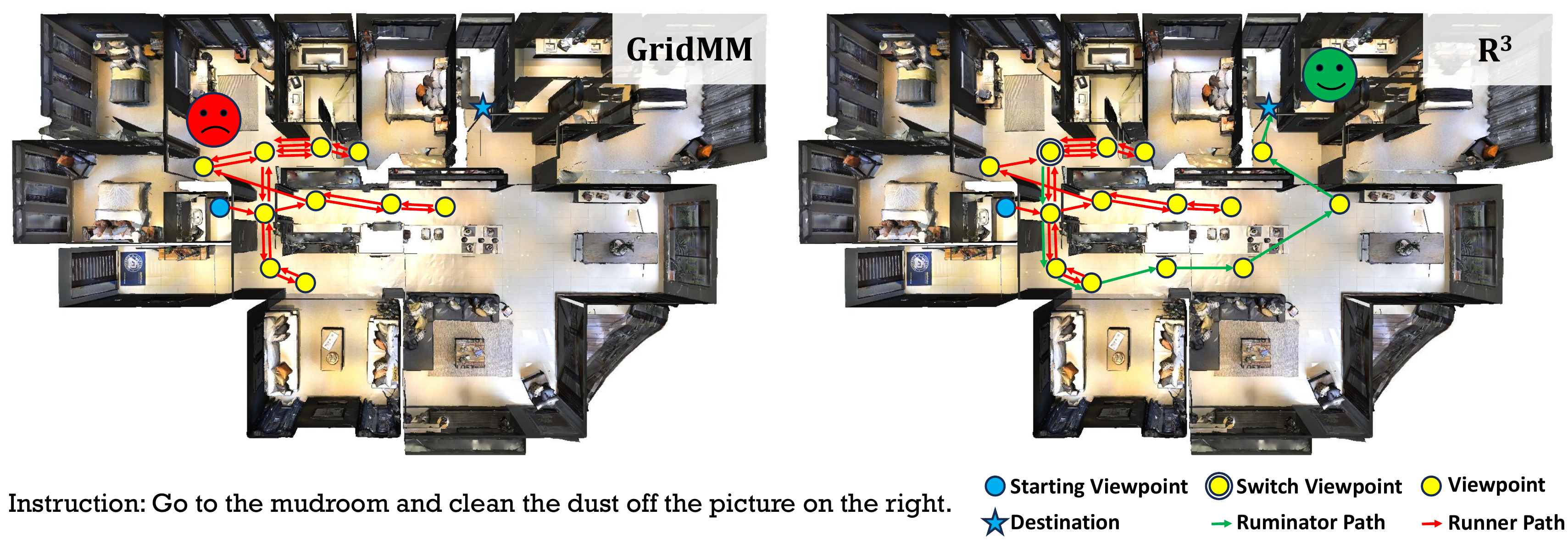}
    \caption{Representative qualitative results on REVERIE validation unseen split.
    Benefiting from long-term planning capabilities, our approach can effectively escape the wandering anomaly and successfully complete the episode.}
    \label{fig:qualitative}
    \vspace{-10pt}
\end{figure*}\\
\textbf{Metrics}.
We utilize standard metrics to measure navigation performance as follows: 
1) Trajectory Length (TL): the navigation path length in meters on average;
(2) Navigation Error (NE): the distance between the agent’s final position and the target in meters on average;
(3) Success Rate (SR): the proportion of trajectories that successfully reach the destination with NE less than 3 meters;
(4) Success weighted by Path Length (SPL): the success rate normalized by the ratio between the length of the shortest path and the predicted path, balancing both SR and TL;
Among these, SPL is widely regarded as the primary measure of navigation performance.
Moreover, we also adopt the following metrics to evaluate the object grounding task:
Remote Grounding Success (RGS): the proportion of tasks that successfully locate the target object (IoU between the predicted bounding box and the ground truth is larger than $0.5$);
and Remote Grounding Success weighted by Path Length (RGSPL): RGS normalized by the path length.
Similar to SPL, RGSPL is regarded as more reflective than RGS.\\
\textbf{Implementation}.
All experiments are conducted on an Ubuntu 16.04.7 LTS server, utilizing Python 3.8.0, PyTorch 1.12.0, and NVIDIA Tesla A100 GPUs.
For the Runner module, we follow the implementation in the official repository of GridMM\citep{GridMM}.
For the Ruminator module, we adopt GPT-4o as the LLM through OpenAI’s official API.
Moreover for the hyperparameters, we set the maximum revisit times $\tau_r=4$, maximum trajectory length $\tau_l=20$, and the scoring threshold $\tau_g=0.35$.

\subsection{Main Results}
In this subsection, we present both qualitative and quantitative experimental comparisons between our R$^3$ and other state-of-the-art methods.

Table~\ref{tab:main} presents comparative results on R2R and REVERIE datasets.
On the R2R validation unseen split, our method outperforms others on all metrics, yielding improvements of 2\% and 1.5\% in terms of SR and SPL.
The increments in performance demonstrate that our method effectively enhances the generalization of VLN agents to unseen environments.
To be noted, our approach significantly exceeds other LLM-based methods, which shows that our exploration makes a solid step toward developing LLMs in the VLN field.
As for the REVERIE validation unseen split, R$^3$ outperforms the best previous methods by a significant margin of 3.28\% and 3.30\% in primary metrics SPL and RGSPL respectively.
We observe that overall elevations on the REVERIE dataset are far more salient than the R2R, exhibiting that our R$^3$ is especially superior for complex tasks where high-level semantic understanding and deliberate analysis are demanded for agents.

Fig.~\ref{fig:qualitative} visualizes trajectories predicted by our approach compared to the SOTA method GridMM.
Although GridMM can initially determine the correct direction toward the destination, it becomes insufficient to capture the global environment layout as exploration expands, leading to wandering around the starting area until navigation fails.
In contrast, triggered by path redundancy, our approach triggers the Ruminator module to effectively recognize the right track to the destination through commonsense reasoning and deliberate analysis of accumulated historical information.

\subsection{Ablation Study}
In this subsection, we carry out ablation studies to further analyze the effect of each component of our approach.
All ablation experiments are conducted on the REVERIE benchmark.

\noindent \textbf{Ablation Study on Regulator}.
Table~\ref{tab:ablation_regulator} validates the effectiveness of each design within the Regulator module.
In the critical evaluation stage, removing any criterion is detrimental to both navigation and object grounding performance, confirming their complementary roles in differentiating anomalies.
Moreover, omitting the \textit{scoring} criterion causes the greatest degradation in navigation performance, leading to 2.05\% and 2.76\% decreases in SR and SPL respectively.
This reveals that the enhancement on navigation metrics is mainly attributed to the \textit{scoring} criterion, which effectively captures the underlying topological structures of navigation history relying on a GNN-based model.
On the other hand, removing the \textit{ending} criterion weakens the object grounding ability most significantly, resulting in declines of 2.33\% and 4.01\% in RGS and RGSPL respectively.
We observe that sometimes the Runner fails to localize the correct object even though having reached the destination.
Under these circumstances, the LLM’s high-level semantic understanding and advanced reasoning ability enable it to realize the mismatch between the localized object and the given instruction and distinguish the correct one in the neighborhood. 
Finally, in the critical formulation stage, we show that skipping the formulation phase leads to a decline in overall effectiveness, underscoring the necessity of a dedicated step.

\begin{table}[t]
  \centering
  \caption{Ablation study on the Regulator. The numbers in script size indicate the decrease in performance compared with our full method (in the first line).}
  \label{tab:ablation_regulator}
  \resizebox{0.48\textwidth}{!}{%
  \begin{tabular}{l|llll}
    \toprule
     & SR$\uparrow$ & SPL$\uparrow$ & RGS$\uparrow$ & RGSPL$\uparrow$ \\
    
    \rowcolor[gray]{0.9}
    \multicolumn{5}{c}{\emph{Ours}} \\

    R$^3$ 
      & \textbf{53.76} 
      & \textbf{42.14} 
      & \textbf{37.94} 
      & \textbf{29.86} \\

    \rowcolor[gray]{0.9}
    \multicolumn{5}{c}{\emph{Critical Evaluation}} \\

    $w/o$ \textit{looping}
      & 53.39 \scriptsize{$\downarrow$0.37} 
      & 41.43 \scriptsize{$\downarrow$0.71} 
      & 37.67 \scriptsize{$\downarrow$0.27} 
      & 28.24 \scriptsize{$\downarrow$1.62} \\

    $w/o$ \textit{scoring}
      & 51.71 \scriptsize{$\downarrow$2.05} 
      & 39.38 \scriptsize{$\downarrow$2.76} 
      & 36.55 \scriptsize{$\downarrow$1.39} 
      & 27.04 \scriptsize{$\downarrow$2.82} \\

    $w/o$ \textit{ending}
      & 53.54 \scriptsize{$\downarrow$0.22} 
      & 40.53 \scriptsize{$\downarrow$1.61} 
      & 35.61 \scriptsize{$\downarrow$2.33} 
      & 25.85 \scriptsize{$\downarrow$4.01} \\

    \rowcolor[gray]{0.9}
    \multicolumn{5}{c}{\emph{Critical Formulation}} \\

    $w/o$ formulation
      & 53.37 \scriptsize{$\downarrow$0.39} 
      & 41.83 \scriptsize{$\downarrow$0.31} 
      & 37.69 \scriptsize{$\downarrow$0.25} 
      & 29.65 \scriptsize{$\downarrow$0.21} \\
    \bottomrule
  \end{tabular}%
  }
\end{table}
\noindent \textbf{Ablation Study on Ruminator}.
In Table~\ref{tab:ablation_ruminator}, we further delve into the Ruminator module’s dependencies on the LLMs' reasoning capabilities.
We present a comprehensive comparison among various LLMs with different reasoning capacities.
Here $w/o$ LLM represents removing the Ruminator and leaving Runner to navigate alone.
We can draw two crucial conclusions from the ablation of backbones: 
(1) As the commonsense reasoning capabilities of LLMs diminish progressively (GPT-4o $>$ GPT-3.5 Turbo $>>$ MiniGPT-4 ~\cite{minigpt}), the overall performance also deteriorates accordingly, highlighting the effectiveness of reasoning capabilities in our system.
This also indicates R$^3$'s potential for further scaling of the generalizing ability as more powerful LLMs become available in the future.
(2) The setting without the Ruminator outperforms MiniGPT-4 by 1.98\% and 0.74\% on SR and SPL metrics, exhibiting that employing inappropriate LLMs can lead to catastrophic results for VLN tasks since they lack both expertise compared with Runner and desired commonsense knowledge demanded for the Ruminator.
Besides, we also compare our results with removing the shared memory bank.
Under this setting, the Ruminator cannot access the Runner's memory and must accumulate historical information from scratch after switching.
This modification leads to a performance decline, emphasizing the importance of enabling the Ruminator to fully exploit historical information.

\section{Conclusion}
This work presents R$^3$, a zero-shot VLN framework that incorporates LLM-driven commonsense reasoning and domain-specific expertise under the dual-process thinking paradigm.
Our approach comprises three modules: Runner, responsible for ensuring efficient and accurate navigation under nominal conditions, Ruminator, dedicated to resolving anomalous situations with deliberate and strategic reasoning in a zero-shot manner, and Regulator, which monitors the navigation progress and adaptively switches the system to the appropriate thinking mode.
Experimental results demonstrate a clear superiority of R$^3$ over other state-of-the-art methods across various benchmarks with different categories of instructions.
Our method especially excels in handling demanding benchmarks such as REVERIE. 
Moreover, it also significantly outperforms other LLM-based methods in inference efficiency.
We hope our work could offer a pragmatic solution to the research community and highlight a novel path for efficiently harnessing LLMs for challenging VLN tasks.

\begin{table}
  \centering
  \caption{Ablation study on the Ruminator.}
  \label{tab:ablation_ruminator}
  \resizebox{0.48\textwidth}{!}{%
  \begin{tabular}{l|llll}
    \toprule
     & SR$\uparrow$ & SPL$\uparrow$ & RGS$\uparrow$ & RGSPL$\uparrow$ \\

    \rowcolor[gray]{0.9}
    \multicolumn{5}{c}{\emph{Backbone}} \\

     GPT-4o (ours) 
      & \textbf{53.76} 
      & \textbf{42.14} 
      & \textbf{37.94} 
      & \textbf{29.86} \\

      BLIP-2 \& GPT-3.5 Turbo  
      & 52.84  \scriptsize{$\downarrow$0.92} 
      & 40.98  \scriptsize{$\downarrow$1.16} 
      & 36.19  \scriptsize{$\downarrow$1.75} 
      & 29.06  \scriptsize{$\downarrow$0.80} \\

      MiniGPT-4   
      & 49.39  \scriptsize{$\downarrow$4.37} 
      & 35.73  \scriptsize{$\downarrow$6.41} 
      & 33.80  \scriptsize{$\downarrow$4.14} 
      & 24.73  \scriptsize{$\downarrow$5.13} \\

      $w/o$ LLM   
      & 51.37  \scriptsize{$\downarrow$2.39} 
      & 36.47  \scriptsize{$\downarrow$5.67} 
      & 34.57  \scriptsize{$\downarrow$3.37} 
      & 24.56  \scriptsize{$\downarrow$5.30} \\

    \rowcolor[gray]{0.9}
    \multicolumn{5}{c}{\emph{Memory}} \\

      $w/o$ memory bank
      & 52.89  \scriptsize{$\downarrow$0.87} 
      & 40.95  \scriptsize{$\downarrow$1.19} 
      & 36.56  \scriptsize{$\downarrow$1.38} 
      & 28.06  \scriptsize{$\downarrow$1.80} \\
    \bottomrule
  \end{tabular}}
\end{table}

\clearpage
\section*{Acknowledgments}
This work is partially supported by the NSF of China (Grants No.62302481, 62525203, U22A2028, 6240073476), Strategic Priority Research Program of the Chinese Academy of Sciences (Grants No.XDB0660200, XDB0660201, XDB0660202), CAS Project for Young Scientists in Basic Research (YSBR-029) and Youth Innovation Promotion Association CAS.

\end{document}